\documentclass[lettersize,journal]{IEEEtran}
\usepackage{amsmath,amsfonts}
\usepackage{algorithm}
\usepackage{algpseudocode}
\usepackage{array}
\usepackage{textcomp}
\usepackage{stfloats}
\usepackage{url}
\usepackage{verbatim}
\usepackage{graphicx}
\usepackage{cite}
\usepackage{hyperref}
\usepackage{xcolor}
\usepackage{harpoon}
\usepackage{mathtools}
\usepackage{multirow}
\usepackage{enumitem}
\usepackage{float}
\usepackage{colortbl}
\usepackage{diagbox}
\usepackage{subfigure}
\usepackage{booktabs}
\usepackage{threeparttable}  

\begin{document}
\title{Safety-aware Goal-oriented Semantic Sensing, Communication, and Control for Robotics}

\author{Wenchao~Wu$^{\ast}$, Shutong~Chen$^{\ast}$, Wenjie~Liu, Zhibo~Pang, Yansha~Deng, and Robert Schober
\thanks{Wenchao~Wu, Shutong~Chen, Wenjie~Liu, and Yansha~Deng are with the Department of Engineering, King’s College London, U.K. (e-mail: wenchao.wu@kcl.ac.uk; shutong.chen@kcl.ac.uk; wenjie.liu@kcl.ac.uk; yansha.deng@kcl.ac.uk). $^\ast$Wenchao~Wu and Shutong~Chen contributed equally to this work. Zhibo Pang is with the Department of Intelligent Systems, KTH Royal Institute of Technology, 10044 Stockholm, Sweden, and the Department of Automation Technology, ABB Corporate Research Sweden, 72226 V\"{a}ster\aa s, Sweden. (email: zhibo@kth.se). Robert Schober is with the Institute for Digital Communications, FriedrichAlexander-Universität Erlangen-Nürnberg, 91058 Erlangen, Germany (email: robert.schober@fau.de). (Corresponding author: Yansha Deng).
}
}


\maketitle

\begin{abstract}
Wirelessly-connected robotic systems empower robots with real-time intelligence by leveraging remote computing resources for decision-making.
However, the data exchange between robots and edge servers often overwhelms communication links, introducing latency that degrades task performance. 
To tackle this, goal-oriented semantic communication (GSC) has been introduced for wirelessly-connected robotic systems to extract and transmit only goal-relevant semantic representations.
While this improves task effectiveness, it generally overlooks practical safety requirements.
Meanwhile, existing robotics research often treats safety primarily as a control-level problem, without systematically considering safety across sensing, communication, and control
in a closed-loop manner.
To bridge this gap, we investigate how to enable safety-aware goal-oriented semantic (SA-GS) sensing, communication, and control co-design in wirelessly-connected robotic systems, aiming to maximize the robotic task effectiveness subject to practical safety requirements.
We first introduce {an} architecture {for} wirelessly-connected
robotic systems and representative use cases. We then summarize general safety requirements and effectiveness metrics across the use cases. 
Next, we systematically analyze the unique safety and effectiveness challenges in sensing, communication, and control. 
Based on these, we further present potential SA-GS research directions.
Finally, {an Unmanned Aerial Vehicle (UAV) target tracking case study validates that one of the presented SA-GS research directions, i.e., semantic-based C\&C packet execution,} could significantly improve safety rate and tracking success rate by more than 2 times and 4.5 times, respectively.
\end{abstract}


\vspace{-0.1in}
\section{Introduction}
Wirelessly-connected robotic system  has emerged as a new paradigm for enabling intelligent, responsive, and decentralized systems.
However, existing wireless networks cannot meet the stringent timeliness requirements of bidirectional communication between robots and the {edge server}, especially in large-scale deployments.
On the one hand, the uplink must support the real-time transmission of bandwidth-intensive multi-modal sensory data, where limited wireless capacity can degrade the {information} freshness and completeness, leading to delayed or suboptimal decision-making.
On the other hand, the downlink transmission of control and command (C\&C) {signals} demands timely transmission with high update frequency and reliability, where any packet loss or congestion might impair the robot’s {task performance}.

To address this, goal-oriented semantic communication (GSC) has been introduced as a promising communication paradigm \cite{zhou2022task}, which extracts and transmits {the important information rather than raw data with respect to the task.}
For {the} uplink, recent research has expanded GSC design for different sensory data types {and} various tasks \cite{sige,10577270}.
{To preserve key geometric relationships and enable accurate downstream visual question answering,} scene graphs were extracted from Unmanned Aerial Vehicle (UAV)-captured images as semantic representations of the original image in \cite{sige}.
To achieve point cloud-based avatar reconstruction in the Metaverse, avatar skeleton nodes were ranked based on their importance for selective transmission in \cite{10577270}.
For {the} downlink, prior work has utilized GSC to selectively transmit and execute {the} most useful C\&C packets \cite{shutong, wenchao}.
The study {in} \cite{shutong} developed a deep reinforcement learning (DRL) algorithm to perform temporal selection {of} C\&C packets, {discarding} packets that contribute little to the downstream control task.
The authors in \cite{wenchao} introduced the value and age of information to quantify the semantic importance and freshness of C\&C packets and drop less informative packets.

Nevertheless, real-world robotic applications demand more than just task effectiveness, they must also meet stringent safety requirements. 
{Based on} industrial robotics standards \cite{eu, us}, both the US and the EU have established formal regulations that define specific safety requirements for different types of robots operating in various scenarios. 
For instance, in UAV target tracking, precise tracking {requires} the UAV {to} stay close to the target, but maintaining a distance of less than 3 meters significantly increases collision risk.
In addition to published standards, some studies in the robotics community have incorporated safety as part of the optimization objective or explicit constraints. 
For example, collision probability has been widely adopted as an optimization objective alongside shortest path planning in autonomous driving applications \cite{10675394}, while for collaborative robots, {the} internal joint force has been used as a key safety criterion \cite{10758369} to trigger {emergency stops to prevent human injury}.
{{However}, existing works in GSC focused only on optimizing effectiveness metrics, such as throughput \cite{10577270} and mean squared error (MSE) \cite{wenchao}, without realizing that safety should be prioritized.}

{Moreover, existing robotics research often treats safety primarily as a control-level problem, without systematically addressing safety across sensing, communication, and control in a closed-loop manner.
However, these three components are inherently coupled. The safety and effectiveness of control depend on the fidelity of sensory data and the reliability of communication. In turn, control decisions shape subsequent sensing and communication processes, while the type of sensory data and the demands of downstream control tasks jointly affect communication frequency and transmission priorities.
Without considering safety in the tight coupling among these three components, at least one dimension is likely to be inadequately addressed, resulting in suboptimal efficiency and potentially compromised system safety.}

Motivated by the above, this work systematically investigates how to enable {safety-aware goal-oriented semantic (SA-GS)} sensing, communication, and control co-design in wirelessly-connected robotic {systems}, aiming to maximize  task effectiveness while {prioritizing safety.}
To the best of our knowledge, this is the first work to extend communication design beyond task effectiveness by explicitly incorporating practical safety requirements into the communication objective.
Our main contributions include:
\begin{itemize}
    \item {We introduce an architecture for wirelessly-connected robotic systems and representative use cases.}
    \item We summarize the safety requirements and effectiveness metrics {{based on} the representative use cases.}
    \item We identify the key challenges of wirelessly-connected robotic systems {for} sensing, communication and control. To tackle these challenges, we present novel {SA-GS future directions across these three components, aiming to maximize task effectiveness subject to practical safety requirements.}
    \item {To validate the effectiveness of the proposed SA-GS directions, we implement one representative direction, namely semantic-based C\&C packet execution, in a single-UAV target-tracking task.} Compared with baseline, it could improve the safety rate by more than 2 times and tracking success rate by more than 4.5 times, respectively.
\end{itemize}

\renewcommand{\arraystretch}{1}
\begin{table*}[]
\caption{Commonly-used sensors in wirelessly-connected robotic {systems}. }
\begin{tabular}{lllll} 
\toprule[2pt]
\multicolumn{5}{c}{\textbf{Robot Embeded Sensor}}           \\  \toprule[2pt]
\textbf{Sensor Type}            & \textbf{Data Type}             & \textbf{Devices}               & \textbf{Transmission Frequency} & \textbf{Data Load} \\ \toprule[2pt]
Tactile Sensor                  & Haptic                   & Tactile Array Sensor           & 100-1000 Hz                     & Low-Medium         \\ \hline
Torque Sensor                   & Force/Torque              & F/T Sensor (e.g., ATI Mini40)  & 500-2000 Hz                     & Low-Medium         \\ \hline
Joint Encoder                   & Joint Angles/Velocities        & Optical or Magnetic Encoder    & 500-1000 Hz                     & Low                \\ \hline
Inertial Measurement Unit (IMU) & Acceleration, Angular Velocity & {MPU6050 IMU}             & 100-1000 Hz                     & Low-Medium         \\ \hline
Magnetometer                    & Heading                   & {BNO055 Orientation Sensor}               & 10-100 Hz                       & Low                \\ \toprule[2pt]
\multicolumn{5}{c}{\textbf{External Sensor}}                                                                                                             \\ \toprule[2pt]
\textbf{Sensor Type}            & \textbf{Data Type}             & \textbf{Devices}               & \textbf{Transmission Frequency} & \textbf{Data Load} \\ \toprule[2pt]
Depth Camera                    & Depth Image                    & Azure Kinect Sensor            & 10-60 Hz                        & Medium-High        \\ \hline
LiDAR Sensor                    & 3D Point Cloud                 & Velodyne, Ouster LiDAR         & 5-20 Hz                         & High               \\ \hline
Ultrasonic Sensor               & Distance                   & {MaxBotix Ultrasonic Estimator}              & 1-50 Hz                         & Low                \\ \hline
UWB Sensor                      & Radio Signal                   & {DW1000 UWB transceiver}               & 10-100 Hz                       & Low                \\ \hline
Wi-Fi Module                    & Radio Signal                   & {Intel AX200 WiFi adapter}        & 1-10 Hz                         & Low                \\ \hline
Infrared Sensor                 & Thermal Imaging           & FLIR Lepton IR Camera          & 1-60 Hz                         & Medium             \\ \toprule[2pt]
\end{tabular}
\label{tab:sensor}
\end{table*}
\renewcommand{\arraystretch}{1}

\section{System Architecture \& Use Cases}
In this section, we first introduce the general architecture and components of a wirelessly-connected robotic system. Based on this, we present three representative use cases that demonstrate the system’s practical applications.

\begin{figure}[t]
    \centering
   {\includegraphics[width=0.9\linewidth]{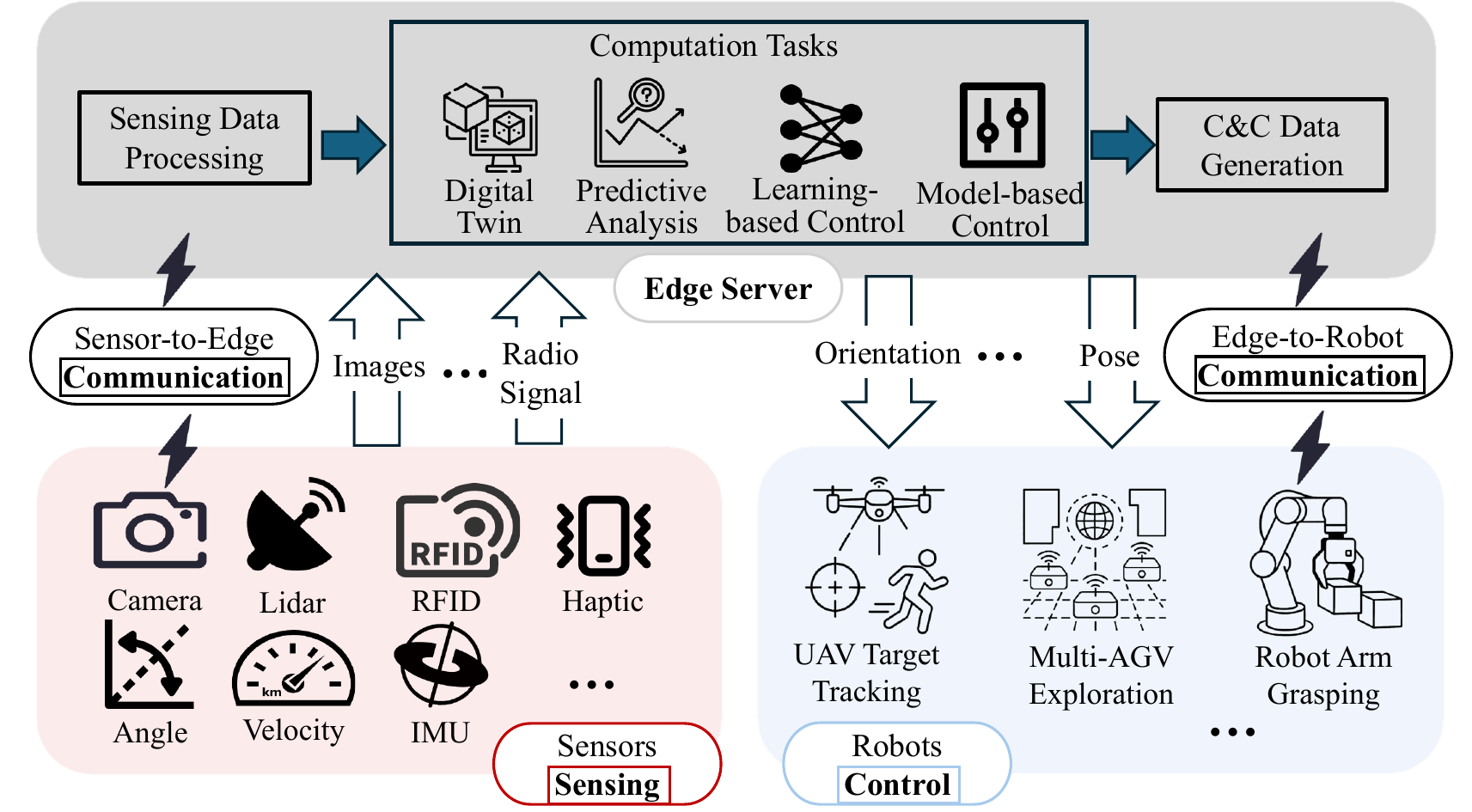}}   
    \caption{{Illustration of a wirelessly-connected robotic system.}}
    \label{fig:system}
    \vspace{-0.2in}
\end{figure}

\subsection{System Architecture}
The wirelessly-connected robotic system consists of five key components: sensors, a sensor-to-edge link, an edge server deployed at the base station, an edge-to-robot link, and a group of robots, as shown in Fig. \ref{fig:system}. 
At the beginning, robot-embedded sensors capture the robot’s internal motion and state, while external sensors collect environmental information, where \textbf{Table~\ref{tab:sensor}} summarizes commonly-used sensors in the system and the data types they collect. 
Subsequently, in the sensor-to-edge link, all the collected sensor data are wirelessly transmitted to the edge server for real-time processing.
Upon receiving these data, the edge server runs real-time simulations (e.g., digital twin) or control algorithms (e.g., learning-based control) to generate optimal control decisions for different industrial robots.
In the edge-to-robot link, these control decisions are translated into executable commands and transmitted back to corresponding robots for action execution.
{To provide a unified basis for the subsequent analysis in Sec. \ref{sec:safety requirements} and \ref{sec:challenges},} we abstract this closed-loop process into three functional stages, namely sensing, communication, and control. 
Specifically, sensing refers to the collection of raw sensor data at the robot side and its processing at the edge server. 
Communication encompasses both sensor-to-edge and edge-to-robot wireless transmissions. 
Control involves generating C\&C packets at the edge server and executing them at the robot side.
\vspace{-0.1in}
\subsection{Use Cases}
{The considered wirelessly-connected robotic system can represent a wide range of robotic applications in practice. Herein, }we present three typical use cases to illustrate its versatility.

\subsubsection{\textbf{Robot Arm Grasping}}
Robot arms can grasp irregular, slippery objects in cluttered environments using multi-modal sensing and edge processing. 
First, images or point clouds are sent to the edge server for object detection and segmentation with vision models like YOLO or PointNet++. After identifying the target, the edge server determines the optimal grasp pose and force using grasp synthesis models such as generative grasping CNN or Contact-GraspNet. 
Based on joint position and velocity data transmitted from the robot’s embedded sensors, the server then plans and returns an efficient, collision-free path. 
During grasp execution, tactile sensors stream feedback for real-time contact stability monitoring, ensuring a secure and reliable grasp.
{However, perception errors or delayed tactile feedback may cause excessive grasping force, posing safety risks to both the object and nearby operators.}
\subsubsection{\textbf{UAV-assisted Tasks}}
UAVs are widely used for parcel delivery and target tracking in industrial settings. 
During operation, vision, positioning, and motion data from cameras, ultrasonic sensors, and IMUs are continuously sent to the edge server for real-time processing. 
Leveraging their top-down view, UAVs use multi-view image fusion to improve tracking accuracy. 
Motion data is also processed by adaptive flight control models at the edge server, enabling dynamic path adjustments in response to obstacles or moving targets.
{However, asynchronous multi-modal data fusion might prevent the UAV from maintaining a safe operating speed, thereby increasing the risk of collisions.}

\subsubsection{\textbf{Multi-robot Exploration}}
Multiple mobile robots often collaborate to map unexplored regions, transmitting diverse sensor data to {the} edge server for localization, motion tracking, and environmental mapping. 
For localization, robots {can} use ultra-wideband (UWB) sensors for time-of-flight measurements or WiFi modules for received signal strength indicator (RSSI)-based distance estimation. 
For motion tracking,  acceleration and velocity data captured by {an IMU can be} fused with radio-based localization data to predict trajectories and prevent collisions. 
For mapping, vision-equipped robots {can} detect landmarks to construct detailed maps, while non-vision robots infer spatial layouts via analyzing the RSSI from multiple fixed RFID tags. 
In the edge-to-robot link, the edge server {can} plan exploration strategies and make coordinated decisions via methods such as cooperative multi-agent DRL (MADRL) to optimize task allocation and exploration efficiency.
{However, delayed localization updates or uncoordinated decision-making may reduce safe inter-robot distance and lead to collisions in dense environments.}

{Although the above use cases share the same architecture, they differ substantially in sensing modalities, task objectives, and decision loops. 
Thus, the associated safety requirements and effectiveness metrics are also task-dependent, {as detailed} in Sec. \ref{sec:safety requirements}.}
\begin{table*}[!ht]
\center
\caption{Table of Effectiveness Metrics}
\vspace{-0.1in}
\resizebox{\linewidth}{!}{ 
\begin{tabular}{ccc} \toprule[2pt]

\textbf{Use Case} &\textbf{Effectiveness Metrics} &\textbf{Description}\\ \midrule[2pt]

\multirow{3}{*}{Robot Arm Grasping} & Grasp Success Rate & The percentage of successful grasps out of the total number of grasping attempts. \\ 
                                    & Execution Time & The time for the robot arm to complete the entire grasping process. \\ 
                                  & Mean Picks Per Hour (MPPH) & The average number of successful picks a robot arm can perform in an hour. \\ \hline
\multirow{6}{*}{UAV-assisted Task} & Tracking Accuracy & Measures how closely the UAV maintains its states relative to targets’ states. \\ 
                                    & Tracking Success Rate & Measures how consistently the UAV can successfully track targets. \\ 
                                  & Identity F1 score (IDF1) & Evaluate performance in maintaining the correct identities of tracked objects over time. \\
                                  & Delivery Accuracy & Measure the deviation between the intended and actual delivery point. \\
                                  & Delivery Success Rate & The proportion of delivery attempts that meets their intended outcomes. \\
                                  & Delivery Time & The total amount of time for a UAV to complete a delivery task. \\ \hline
\multirow{2}{*}{Multi-robot Exploration} & Mapping Time & The time to explore 95\% of the target area.\\ 
                                    & Mapping Error & Discrepancies in the maps generated by various robots during exploration.

 \\ \bottomrule[2pt]
\end{tabular}
}	

\label{Effectiveness Metrics}
\vspace{-0.2in}
\end{table*}
\begin{table*}[!ht]
\center
\caption{Types of Semantic Representations.}
\vspace{-0.2in}
\resizebox{\linewidth}{!}{ 
\begin{tabular}{c c c c c c c c} \toprule[2pt]

\multirow{2}{*}{\textbf{Semantic Representations}} &\multirow{2}{*}{\textbf{Data Type}} &\multicolumn{2}{c}{\textbf{Robot Arm Grasping}}&\multicolumn{2}{c}{\textbf{UAV-assisted Task}}&\multicolumn{2}{c}{\textbf{Multi-robot Exploration}}\\ \cmidrule[2pt]{3-8}

\multirow{2}{*}{}&\multirow{2}{*}{} &\textbf{Safety}&\textbf{Effectiveness} &\textbf{Safety}&\textbf{Effectiveness} &\textbf{Safety}&\textbf{Effectiveness}   \\ \midrule[2pt]

\multirow{1}{*}{Object’s attributes
(e.g., type, material, size, and surface texture)
}&\multirow{1}{*}{Image, Point cloud} &\checkmark&	\checkmark	& {$\mathrm{o}$}	&\checkmark&$-$	& $-$ \\ \hline

\multirow{1}{*}{Object 6D pose}&\multirow{1}{*}{Point cloud} &\checkmark&	\checkmark	&{$\mathrm{o}$}	&\checkmark&$-$	&$-$	  \\ \hline

\multirow{1}{*}{Grasping position}&\multirow{1}{*}{Point cloud} &\checkmark&	\checkmark	&$-$	&$-$&$-$	&$-$	  \\ \hline

\multirow{1}{*}{Obstacles (e.g., walls, roads, and barriers)}&\multirow{1}{*}{Image, Point cloud} &\checkmark&	\checkmark	&\checkmark	&{$\mathrm{o}$}&	\checkmark&	\checkmark  \\ \hline

\multirow{1}{*}{Obstacle’s 6D pose}&\multirow{1}{*}{Point cloud} &\checkmark&	\checkmark	&\checkmark	&{$\mathrm{o}$}&	\checkmark&	\checkmark  \\ \hline

\multirow{1}{*}{Number of surrounding objects}&\multirow{1}{*}{Image, Point cloud} &\checkmark&	\checkmark	&\checkmark	&{$\mathrm{o}$}&	\checkmark&	\checkmark  \\ \hline

\multirow{1}{*}{Relative distance to obstacles}&\multirow{1}{*}{Image, Point cloud} &\checkmark&	{$\mathrm{o}$}	&\checkmark	&{$\mathrm{o}$}&	\checkmark&	\checkmark  \\ \hline

\multirow{1}{*}{Relative speed to obstacles}&\multirow{1}{*}{Video} &\checkmark&	{$\mathrm{o}$}	&\checkmark	&{$\mathrm{o}$}&	\checkmark&	\checkmark  \\ \hline

\multirow{1}{*}{Holding or not}&\multirow{1}{*}{Image, Point Cloud} &\checkmark&	\checkmark	&$-$	&$-$&$-$	&$-$	  \\ \hline

\multirow{1}{*}{Object deformation}&\multirow{1}{*}{Point Cloud} &\checkmark&	{$\mathrm{o}$}	&$-$	&$-$&$-$	&$-$	  \\ \hline

\multirow{1}{*}{Object texture, Grasping force
}&\multirow{1}{*}{Haptic} &\checkmark&	\checkmark	&$-$	&$-$&$-$	&$-$	  \\ \hline

\multirow{1}{*}{Robots’ locations
}&\multirow{1}{*}{3D radio signal} &$-$&$-$		&\checkmark	&\checkmark&	\checkmark&	\checkmark  \\ \hline

\multirow{1}{*}{Relative distance between robots
}&\multirow{1}{*}{Image, Point cloud} &$-$&$-$		&$-$	&$-$&	\checkmark&	\checkmark  \\ \hline

\multirow{1}{*}{Relative speed between robots
}&\multirow{1}{*}{Video} &$-$&	$-$	&$-$	&$-$&	\checkmark&	\checkmark  \\ \bottomrule[2pt]
\end{tabular}
}	

\label{Semantic Representations}
\begin{tablenotes}
        \footnotesize
        \item \textit{Note:} ``$\checkmark$", ``{$\mathrm{o}$}", and ``$-$" means {compulsory, optional}, and not applicable for safety or effectiveness in the use case, respectively. {Here, compulsory semantic representations are those whose absence may directly violate a safety constraint or harm the downstream task performance, whereas optional semantic representations are not strictly required for but can further improve safe and effective operation.}
\end{tablenotes}
\vspace{-0.2in}
\end{table*}

\section{Safety Requirements \& Effectiveness Metrics}
\label{sec:safety requirements}
In this section, safety requirements and effectiveness metrics are defined to evaluate the performance of wireless-connected {robotic systems for} the given use cases. 
\vspace{-0.2in}
\subsection{Safety Requirements}
To ensure the safe operation of {a} robotic system, the following requirements need to be satisfied.
\subsubsection{\textbf{Safety Distance}}
{This} refers to the minimum separation required between robots and objects to ensure safety. 
{The safety distance} can be measured using {the concept of} danger field, including static danger field (SDF) and kineto-static danger field (KSDF) \cite{Safety_related_distance_1}.
Moreover, it can be quantified by maintaining minimum distance guarantees, which may be fixed at all times or dynamic based on {the} real-time relative motion and velocity between robots and objects \cite{eu}. 

\subsubsection{\textbf{Safety Tracking Error}}
{This} is the minimum deviation required between the robot's intended and actual trajectories to prevent collisions with the surrounding environment. 
{It} can be quantified by the real-time average distance between the robot's target and actual positions \cite{us}, or the lateral offset and yaw angle error \cite{Safety_related_tracking_error_2}.

\subsubsection{\textbf{Safety Grasping Force}}
{This} represents the required force range applied by a robotic gripper/manipulator to safely handle objects without causing damage. 
{It} involves defining upper and lower grasping force thresholds, which can be quantified by building a contact force model related to {the} stiffness, friction{,} and weight of objects, such as the single point with friction (SPwF) contact model \cite{Safety_related_grasping_force_1}.

\subsubsection{\textbf{Safety Speed}}
{This} refers to the maximum operating speed that ensures safety. 
For robot arms, ISO/TS 15066 and ISO 10218 specify that the speed of an end effector should be limited to 250 mm/s when a human enters the workspace.
For mobile robots, ISO 3691-4 specifies speed limits. When obstacles are within 500 mm laterally, the maximum speed is 1.2 m/s. 
If {additionally} frontal clearance is $\leq$ 500 mm, the limit is reduced to 0.7 m/s. Robots lacking personnel detection are restricted to 0.3 m/s.
\subsubsection{\textbf{Quasi-dynamic Stability}}
This refers to a legged robot’s ability to maintain balance during walking, running, or jumping with payloads, thereby avoiding risks to nearby equipment and personnel. 
It is quantified using metrics such as the Zero Moment Point (ZMP) \cite{stability}, Center of Pressure (CoP), and Force Restoration Index (FRI), which assess whether {the} torques and ground reaction forces {experienced} at the contact points remain within the support polygon to prevent tipping during dynamic motion.
\vspace{-0.2in}
\subsection{Effectiveness Metrics}
Effectiveness metrics vary by {task} and are summarized for the given use cases in \textbf{Table \ref{Effectiveness Metrics}}. 
For robot arm grasping, key metrics include \textit{\textbf{Grasp Success Rate}}, \textit{\textbf{Execution Time}}, and \textit{\textbf{Mean Picks Per Hour (MPPH)}} \cite{MPPH}. 
{For} UAV-assisted tasks, two scenarios are considered: target tracking and parcel delivery. 
For target tracking, effectiveness is measured by \textit{\textbf{Tracking Accuracy}}, \textit{\textbf{Tracking Success Rate}}, and \textit{\textbf{Identity F1 Score (IDF1)}} \cite{IDF1}. 
For parcel delivery, effectiveness metrics include \textit{\textbf{Delivery Accuracy}}, \textit{\textbf{Delivery Success Rate}}, and \textit{\textbf{Delivery Time}}. 
Lastly, for multi-robot exploration, performance is evaluated based on \textit{\textbf{Mapping Time}} and \textit{\textbf{Mapping Error}}.

\vspace{-0.1in}
\section{Challenges \& {Future Research Directions}}
\label{sec:challenges}
In this section, we examine the key challenges in sensing, communication, and control from the perspective of improving task effectiveness while prioritizing safety. 
The main difficulty is that, under limited sensing, communication, and computation resources, robotic systems cannot maintain all information with equal fidelity, timeliness, and reliability. 
This gives rise to a fundamental trade-off: {safety-critical information should be prioritized in sensing and transmission, and safety-critical commands should be executed with precedence}, while the remaining resources can be used to enhance task effectiveness. 
{Motivated by this observation, we discuss the major challenges across these three components and outline corresponding potential SA-GS research directions, which could be validated in future work.}
In the following discussion, we denote the key challenges by ``C" and the potential research directions by ``D" for clarity.
\vspace{-0.2in}



\subsection{Sensing}
Sensing is the process of robotic systems to perceive {their environment}, but several challenges limit the ability to ensure safety and task effectiveness. 
At the object level, insufficient semantic {representation} extraction, difficulties in point cloud segmentation, and inefficient object recognition for unknown objects and aerial views, compromise perception accuracy. At the environmental level, high computational costs in multi-sensor data processing and asynchronous multi-modal sensor data degrade situational awareness and hinder timely, safety-critical decisions. The following subsections detail these challenges and potential research directions. 

\textbf{C1: \textit{Insufficient Semantic Representations Extraction.}} Robots need to extract information from sensor data to recognize their surrounding objects. However, current models only extract limited information (e.g., object type), resulting in inefficient object recognition, compromising both safety and effectiveness. For instance, in real-time path planning, ignoring critical information like relative distance between robots and obstacles can lead to unsafe control decisions, violating {the} safety distance, thereby increasing collision risk and mapping time. 

\textbf{{D1}: \textit{Semantic Representations Definition.}} To address this challenge, we define semantic representations that quantify the importance for task safety and effectiveness. As safety requirements and effectiveness metrics vary by tasks, these representations are tailored to each context. \textbf{Table \ref{Semantic Representations}} summarizes them across {the} three given use cases, highlighting how they differ based on task-specific requirements. 

\textbf{C2: \textit{Point Cloud Segmentation.}} Point cloud data is inherently sparse and irregular, making accurate object segmentation challenging, especially when objects blend with geometrically similar backgrounds. This can cause merged boundaries and obscure features, leading to missed obstacle detections during navigation or inaccurate grasp planning, which results in safety risks and reduces effectiveness. 

\textbf{{D2}: \textit{Semantic-based Segmentation.}} To solve this, the state-of-the-art approach uses contrastive learning to sub-sample and group same-class points near each boundary, followed by average pooling and class distribution analysis for multi-scale refinement. However, this is computationally expensive, as all objects are treated equally regardless of their importance to the task. {A promising future direction is to propose a more efficient semantic-based approach}, which detects important objects and prioritizes boundary segmentation for them, reducing computation while preserving safety and effectiveness.

\textbf{C3: \textit{Inefficient Object Recognition.}} 
Current object recognition models are inefficient, particularly for unknown objects or those imaged from varying altitudes and angles (e.g., top-down), which may cause distortion and misclassifications. Such errors can violate safety requirements on grasping force, leading to slippage and increased execution time. To address these, {we outline the following future research directions}.

\textbf{{D3.1}: \textit{Semantic-based Unknown Object Recognition.}} Existing methods for unknown object recognition largely adopt open-set detection, using uncertainty estimation or knowledge transfer to distinguish known from unknown classes. While effective for classification, they cannot ensure safe, reliable task execution. Thus, {One promising direction is to emphasize semantic representations (e.g., texture, material)}. This can also be extended to unknown-object grasping, where current methods struggle with generalization and sim-to-real transfer. Semantic data augmentation can enhance policy robustness, while domain adaptation and feature fusion can improve alignment between simulated and real-world distributions.

\textbf{{D3.2}: \textit{Multi-view Image Fusion.}} UAV aerial images are typically captured from bird’s-eye views, rendering side-view–oriented 3D object detection models ineffective. A common solution is monocular 3D detection with height offsets, but it is highly sensitive to height errors and viewpoint distortions. {Multi-view image fusion could serve as a potential direction by capturing objects from multiple angles and enforcing: (1) projection consistency, aligning 2D projections of predicted 3D boxes with ground truth; and (2) multi-view consistency, ensuring stable 3D boxes across views.} This could reduce reliance on height estimates and mitigates viewpoint distortion.

\textbf{C4: \textit{Multiple Sensor Data Processing.}}
Robots transmit large volumes of sensor data for edge processing, but limited computational resources can cause delays that compromise safety and effectiveness. In multi-robot exploration, local images are queued for processing. Delays in critical images' (e.g., robots nearing obstacles) processing may violate safety distance requirements, leading to collisions and longer mapping times. To address this, we propose several {research directions}.

\textbf{{D4.1}: \textit{Semantic-based Queue Priority Design.}} Traditional queuing methods, such as first-in–first-out (FIFO) and first-in–last-out, prioritize data solely by arrival time, which can delay processing of critical sensor data. For example, in FIFO, late-arriving but safety-critical data may be processed too late. Thus, {a promising solution is to design a semantic-based queue priority scheme, which ranks data by its importance.} Specifically, more important sensor data (e.g., images with numerous obstacles) is prioritized for earlier processing.

\textbf{{D4.2}: \textit{Semantic-based Task Offloading.}} Offloading sensor data processing from edge to cloud can reduce computational delays, but existing methods rely solely on bit-level metrics, ignoring semantic representations. This may lead to inefficient offloading, such as processing non-critical images (e.g., static robots) while delaying critical ones (e.g., robots near obstacles), increasing collision risk. {Thus, one direction is to design a semantic-aware offloading approach that offloads sensor data processing based on its importance to safety and effectiveness.}


\textbf{C5: \textit{Sensor Data Synchronization.}} The edge server receives multi-modal data from heterogeneous sensors with different sampling rates, often leading to asynchronous arrivals and misaligned sensor fusion. For instance, misalignment between tactile signals and point-cloud data during grasping can yield inaccurate deformation estimates, leading to unsafe forces and reduced MPPH. This requires a synchronization method that aligns incoming data according to its temporal resolutions.

\textbf{{D5}: \textit{Multi-sensor Data Synchronization.}} Existing synchronization methods include software-based schemes, such as network time protocol (NTP), and hardware-based schemes, such as IEEE 1588-2008 precision time protocol (PTP). While both can achieve sub-microsecond precision, NTP suffers from limited accuracy in dynamic networks, and PTP requires specialized hardware, limiting deployment. {As not all data demands equal precision, one direction is to propose a hierarchical framework to allocate higher precision to safety/effectiveness-critical data (e.g., tactile signals) and lower precision to others.}

\subsection{Communication}
Communication refers to the process of transmitting both sensor and C\&C packets, with the following challenges and potential research directions.

\textbf{C6: \textit{Limited Communication Resources.}} Limited communication resources for uplink sensor and downlink C\&C packets transmission could cause delays/distortions that compromise safety and efficiency, especially for dense point cloud data (e.g., Ford 01 dataset: 120 million points over 1500 frames at 54 bits/point). This may prevent timely obstacle recognition, increasing collision risks and mapping time. To address this, we {outline} the following potential research directions.

\textbf{{D6.1}: \textit{Semantic-based Sensor Data Compression.}} One approach is to reduce sensor data size via compression. However, existing methods focus on full data reconstruction rather than selectively preserving semantic representations, leading to inefficiency. Thus, we {outline} the following potential {research directions}:
\begin{itemize}
    \item \textbf{\textit{Semantic-based Traditional Compression Redesign:}} Traditional compression applies uniform strategies across data, yielding consistent ratios regardless of each part’s importance to safety and effectiveness. {Thus, a semantic-based redesign could be outlined as a direction, which guides compression by importance. Specifically, in transform-based compression (e.g., JPEG), semantic importance can guide distortion weighting and bit allocation across blocks or coefficients. In octree-based point-cloud compression, partitioning depth and point retention can be adaptively adjusted according to the semantic importance of local regions. In graph-based compression, node or edge preservation can be prioritized based on their contribution to downstream tasks.}
    \item \textbf{\textit{Semantic-based Rate-distortion Optimization:}} Learned compression methods typically optimize bit-level rate–distortion functions, which may be too stringent for safe/effective task execution and lead to suboptimal compression. For example, a method that retains only target objects while discarding the background may enable effective grasping but is penalized under conventional rate–distortion metrics due to excessive image distortion. {To address this, one direction is to design semantic-based rate–distortion functions, which can replace bit rate with \textit{semantic representation rate} (semantic content transmitted per time slot) and distortion with \textit{semantic representation distortion} (deviation in reconstructed semantics).}
\end{itemize}

\textbf{{D6.2}: \textit{Semantic-aware Resource Allocation and Scheduling.}} Another research direction is to jointly optimize resource allocation and scheduling for sensor or C\&C packets transmission. Existing schemes, such as proportional fair scheduling and frequency-domain packet scheduling (FDPS), focus solely on bit-level uplink performance, resulting in inefficient resource utilization. For example, FDPS may allocate minimal resources to critical data (e.g., images containing target objects) under poor channel conditions, delaying transmission. This highlights the need for semantic-aware resource allocation and scheduling, where more communication resources are allocated to more important sensor or C\&C packets to ensure timely and reliable transmission.

\textbf{{D6.3}: \textit{Semantic-aware Dynamic Transmission.}} Existing fixed-period transmission of sensor or C\&C packets wastes resources on low-importance data. This {introduces another direction} to design a semantic-aware dynamic transmission approach, which adapts transmission rate based on data importance. Specifically, only essential data is transmitted, with rates increased in high-risk scenarios (e.g., dense obstacles) or under more stringent safety requirements, enabling more frequent delivery of critical data.

\subsection{Control}
Control involves generating high-level decisions and executing low-level commands, but several challenges hinder its ability to guarantee safety and task effectiveness. At the higher level, complex coordinated decision-making and unstable real-time path replanning can result in unsafe or ineffective actions. At the lower level, incorrect C\&C packets may control the robot into unsafe states. The following subsections outline these challenges and potential research directions.

\textbf{C7: \textit{Coordinated Decision-making.}} Effective multi-robot collaboration requires real-time coordinated decision-making, which becomes more challenging as robot number grows. This highlights the need for a coordinated decision-making strategy that dynamically adapts to task progress and inter-robot dependencies, ensuring both safety and effectiveness.

\textbf{{D7}: \textit{Multi-robot Collaboration.}} Two common coordinated decision-making strategies are game theory and cooperative MADRL. Game theory yields stable but suboptimal strategies by balancing each robot’s benefits (e.g., area explored), costs (e.g., travel distance), and safety (e.g., collision risk). Cooperative MADRL maximizes shared global performance but potentially inducing unsafe individual actions. A promising {direction} is to combine decentralized large language models (LLMs) with MPC. Each robot uses an LLM to exchange capabilities and intentions, collaboratively generating a high-level cooperation plan. MPC then translates this plan into low-level motion commands and validates safety via future-state simulations before execution.

\textbf{C8: \textit{Path-replanning Algorithms Design.}} In real-world environments, robots must replan trajectories in real time to avoid sudden obstacles. Excessive replanning can cause motion instability, whereas delayed responses increase collision risk. A novel path-replanning algorithm is needed to balance responsiveness and stability, ensuring safe and effective adaptation to dynamic changes.

\textbf{{D8}: \textit{Fast and Adaptive Replanning.}} 
UAVs and mobile robots often employ DRL-based replanners to decide when to replan, but lack dynamic control over replanning frequency. Vision-based methods with vision–language models are commonly used in robot arms to infer spatial relationships (e.g., obstacle proximity), but slow processing can yield outdated decisions. Thus, a promising {direction} is to design a hierarchical replanning framework: 1) a fast, low-level replanner ensures real-time safety via short-horizon control, triggered when safety is at risk; and 2) a slower, high-level replanner refines the global trajectory to optimize long-term effectiveness.

\textbf{C9: \textit{C\&C {Packet} Execution.}} Safe and effective task execution requires robots to {receive correct} C\&C packets, as {erroneous} C\&C packets can hinder the robot's ability to maintain the safety distance requirement, risking damage to the robot. This highlights the need to ensure optimal C\&C execution.

\textbf{{D9}: \textit{Semantic-based C\&C Packet Execution.}} Similar to sensor data, incoming C\&C packets are queued for execution in robots. Existing methods execute only the latest data, which can be unsafe if it is delayed or outdated. {Thus, one direction is to design a semantic-based execution strategy that could prioritize C\&C packets by their importance to task safety and effectiveness.} The importance could be ranked by freshness, quantified via age of information, and data value, measured by value of information.

\section{Case Study}
{To validate the effectiveness of the proposed SA-GS directions, we implement one representative direction, namely semantic-based C\&C packet execution (D9), referred to as SemCE, in a single-UAV target-tracking task. 
Specifically, an edge server periodically transmits C\&C data to control a UAV to track a mobile target (eg., vehicle) moving over $T$ time slots.
The mobile target moves along a randomly generated trajectory, with a constant speed within each time slot that may vary across time slots.
We assume the mobile target can send its real-time positions to the edge server correctly all the time without error.
Meanwhile, the UAV's onboard sensor captures its real-time positions, also forwarding them to the edge server.
At each time slot $t\in\{1,2,\ldots,T\}$, the edge server generates a C\&C packet according to the current tracking state using a proportional integral derivative (PID) controller with proportional gain 0.5. 
The C\&C packet is transmitted through a Rayleigh-fading channel.
As a baseline, the UAV executes the most recently received C\&C packet, referred to as the latest-arrived execution policy. 
By contrast, SemCE prioritizes received packets based on their freshness and value of information.
Let $d_t$ denote the Euclidean distance between the UAV and the target. Let $d_\mathrm{s}$ and $d_\mathrm{e}$ denote the safety distance and tracking threshold, respectively, where $d_\mathrm{e}>d_\mathrm{s}>0$. Then, the tracking status at time slot $t$ is classified as unsafe if $d_t\leq d_\mathrm{s}$, successful if $d_\mathrm{s}<d_t\leq d_\mathrm{e}$, and unsuccessful if $d_t>d_\mathrm{e}$. 
Accordingly, safety and effectiveness are evaluated by the safety rate and tracking success rate, respectively. 
The safety rate is defined as $T_\mathrm{s}/T$, where $T_\mathrm{s}$ is the number of safe time slots, while the tracking success rate is the proportion of time slots satisfying $d_\mathrm{s}<d_t\leq d_\mathrm{e}$.}
\begin{figure}[htbp]
    \centering
    \subfigure[$d_\mathrm{s}=1 $ (m).]{\includegraphics[width=0.4\linewidth]{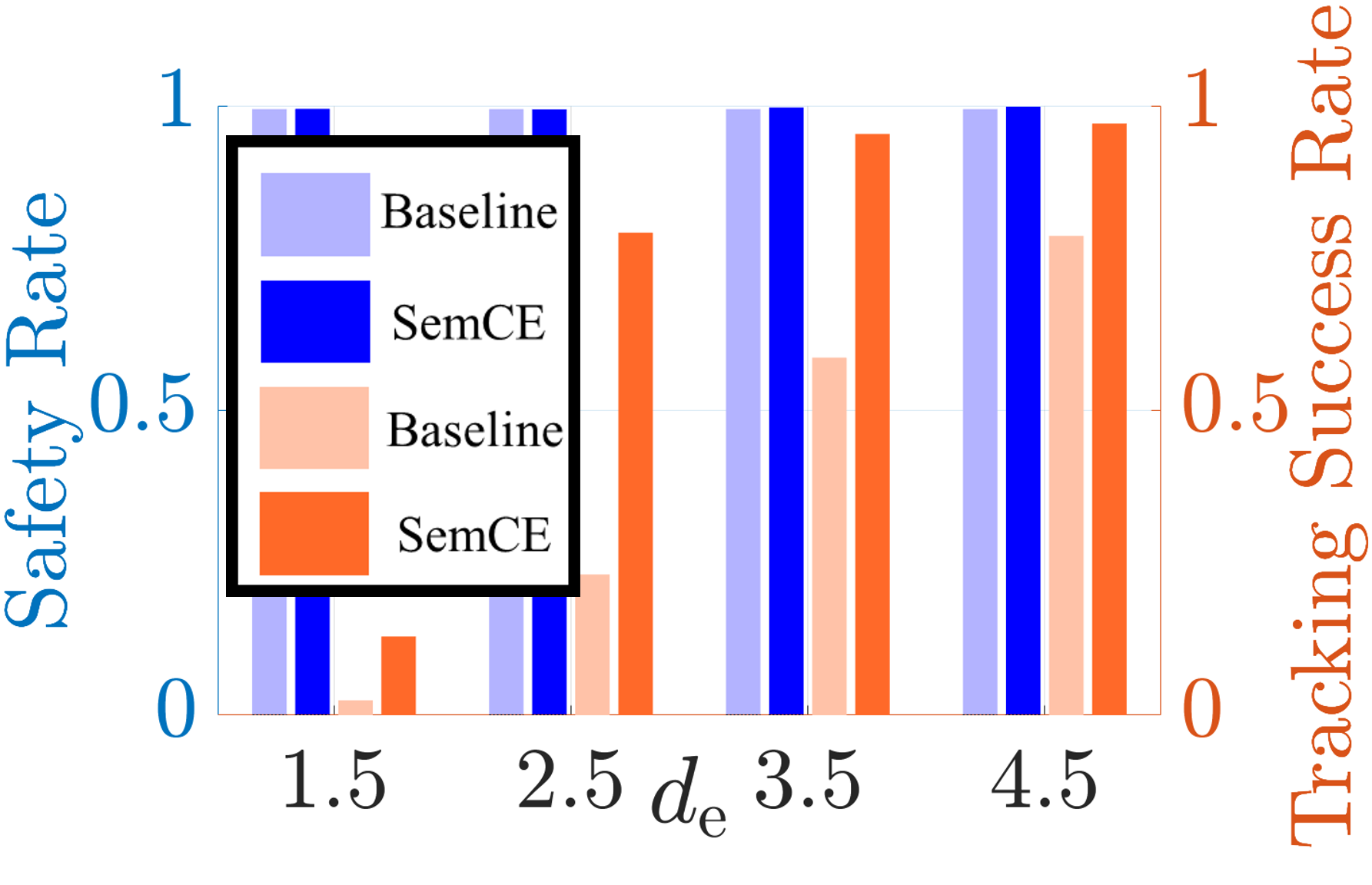}}
    \subfigure[$d_\mathrm{e}=5$ (m).]{\includegraphics[width=0.4\linewidth]{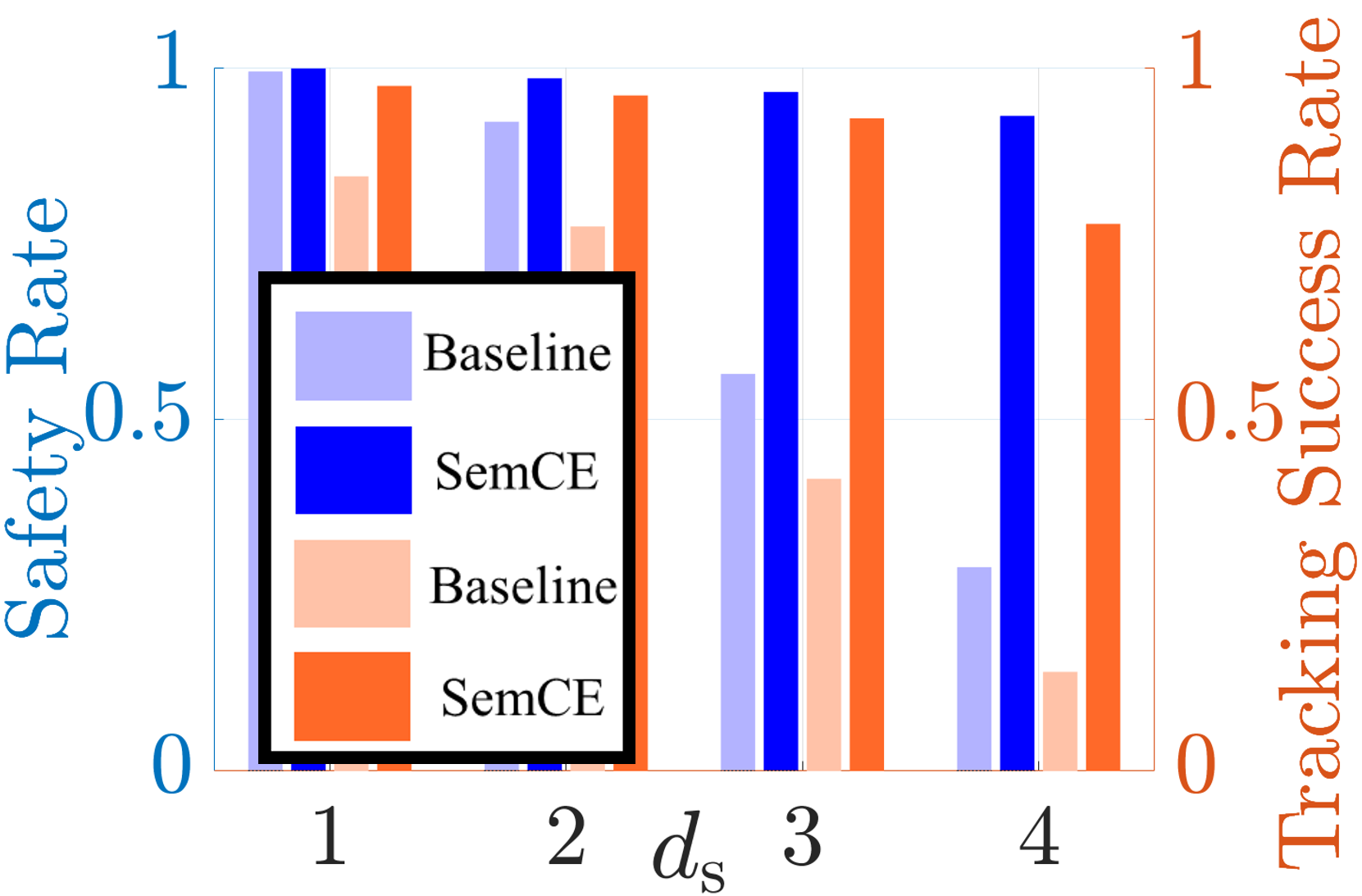}}    
    \caption{Safety and tracking success rate comparisons over various $d_\mathrm{s}$ and $d_\mathrm{e}$.}
    \label{Performance}
\end{figure}
\par {Fig. \ref{Performance} compares the safety rate (blue bars) and tracking success rate (orange bars) of the baseline and SemCE under two threshold settings. 
In Fig. \ref{Performance} (a), the safety distance is fixed at $d_\mathrm{s}=1$ m, while the tracking threshold varies as $d_\mathrm{e}\in\{1.5,2.5,3.5,4.5\}$ m. 
In Fig. \ref{Performance} (b), the tracking threshold is fixed at $d_\mathrm{e}=5$ m, while the safety distance varies as $d_\mathrm{s}\in\{1,2,3,4\}$ m. 
It is observed that SemCE consistently outperforms the latest-arrived baseline in both safety rate and tracking success rate across all evaluated settings. 
In Fig. \ref{Performance} (a), with $d_\mathrm{s}$ fixed at 1 m and varying $d_\mathrm{e}$, both schemes maintain high safety rates as the safety requirement is not stringent. 
As $d_\mathrm{e}$ decreases and approaches $d_\mathrm{s}$, the success rate of both schemes declines due to a narrower tracking range.
In Fig. \ref{Performance} (b), as $d_\mathrm{s}$ increases with fixed $d_\mathrm{e}=5$ m, both safety and tracking become more challenging, leading to performance degradation for both schemes. 
Nevertheless, SemCE exhibits a markedly smaller drop, indicating that prioritizing fresher and more important C\&C packets improves both safe operation and tracking effectiveness. 
}

\section{Conclusion}
{We investigated how to enable safety-aware goal-oriented semantic (SA-GS) sensing, communication, and control co-design in wirelessly-connected robotic systems, with the objective of maximizing task effectiveness subject to practical safety requirements. 
We first introduced the architecture and representative use cases of wirelessly-connected robotic systems. 
We then summarized the general safety requirements and effectiveness metrics across these use cases. 
Next, we systematically analyzed the key safety and effectiveness challenges in sensing, communication, and control. 
Based on this analysis, we presented potential SA-GS directions. Finally, a UAV target-tracking case study demonstrated that these directions can improve the safety rate and tracking success rate by more than 2 times and 4.5 times, respectively.}
\bibliographystyle{IEEEtran}
\bibliography{mylib}

\end{document}